\documentclass[sigconf]{acmart}

\usepackage{booktabs} 
\usepackage{algorithm}  
\usepackage{algorithmicx}  
\usepackage{algpseudocode}
\usepackage{graphicx}

\setcopyright{rightsretained}

\acmDOI{10.475/123_4}

\acmISBN{123-4567-24-567/08/06}

\acmArticle{4}
\acmPrice{15.00}

\begin{document}
\title{Automatic Derivation Of Formulas \\ Using Reforcement Learning}
\titlenote{The Source Code:\url{https://github.com/Luomin1993/SciBot}}

\author{MinZhong Luo}
\affiliation{%
  \institution{China Institute of Atomic Energy}
  \streetaddress{Department of nuclear technology application}
  \city{Peking}
}
\email{luomincong@foxmail.com}

\author{Li Liu}
\affiliation{
  \institution{China Institute of Atomic Energy}
  \streetaddress{Reactor Engineering Technology Research Division}
  \city{Peking}
}
\email{liuli@foxmail.com}

\renewcommand{\shortauthors}{Luo et al.}

\begin{abstract}
This paper presents an artificial intelligence algorithm that can be used to derive formulas from various scientific disciplines called \textbf{automatic derivation machine}. First, the formula is abstractly expressed as a multiway tree model, and then each step of the formula derivation transformation is abstracted as a mapping of multiway trees. Derivation steps similar can be expressed as a reusable formula template by a multiway tree map. After that, the formula multiway tree is eigen-encoded to feature vectors construct the feature space of formulas, the Q-learning model using in this feature space can achieve the derivation by making training data from derivation process. Finally, an automatic formula derivation machine is made to choose the next derivation step based on the current state and object. We also make an example about the nuclear reactor physics problem to show how the automatic derivation machine works.
\end{abstract}

%
%



\maketitle

\section{Introduction}

The automatic derivation of formulas is a revolutionary application of AI in a particular field of scientific research.  By abstracting the structure of a formula into a deducible data structure, the goal can be deduced. Previously there has been no study of formula-derived AI methods, but "automatic machine proof" \cite{Tarski1951A} is a study that is similar to formula derivation. Automatic machine proof refers to the method which transforms proof-type problems into computer-computable forms, such as polynomials \cite{Collins1998Partial}, and then calculating it. Automatic proof is a special form of automatic formula derivation. Due to the requirement of simple calculation, machine proof generally involves only simple geometric problems \cite{Wu1986Basic}.


The automatic formula derivation in this paper refers to the use of the AI method in various professional fields (such as materials science, computational mechanics, etc.) to make complex calculations of relevant formulas automatically to obtain valuable calculation results and solutions, which is revolutionary for traditional scientific research.


First, we re-express the formula as a multiway tree model, which is called a formula multiway tree. In the formula multiway tree, each non-leaf node is an operation symbol (for example, $=$, $+$, $exp(.)$ Etc), while leaf nodes are algebraic and numeric symbol (eg $x$, $e^{-2.337}$, $\pi$, etc.). By devising the subtree matching algorithm,tree construction algorithm,subtree replacement algorithm of the formula, the formula can be deduced or transformed. The subtree matching algorithm and replacement algorithm of the formula multiway tree can make the formula multiway tree deform according to a certain template, which makes the complex formula can be deformed according to the simple formula multiway tree deformation method, and this deformation mode can be used for more complex formulae, which is called iterative learning for automatic formula derivation.


We have designed the encoding algorithm for the formula multiway tree, which can transform different formulas into feature vectors in a unified dimension and establish the feature space of the formulas. The eigencoding of the formula multiway tree makes: 

\begin{itemize}
\item each formula has a unique identity;
\item we can measure the similarity of the formulas. The similar formula multiway trees have a smaller spatial distance;
\item the feature vectors can be used to train the learner.
\end{itemize}


We use the reinforcement learning mechanism to train the formula derivation machine. For a specific professional problem (such as calculation in theoretical mechanics), we need to prepare the relevant derivation training set, set the optional formula multiway tree transformation method to the action set ${a_i}$, and set expressions for formulas at different stages to state set ${s_i}$, and use the gradient descent method to optimize the model parameters by establishing the neural network model $\pi_{\theta}(a|s)$.


There are two main difficulties in the automatic derivation of formulas.

(1) The formula itself is a highly abstract mathematical language and cannot be directly calculated like other training data for machine learning. 

(2) There are many basic formulas in each professional field. Man-made annotations have high professional requirements for people. For the first difficulty, the formula multiway tree can decompose the formula to the calculation-symbol level, can contain all the information of a formula, and its encoding can also be used as a training data to input learner. For the second difficulty, we can use the template mapping mechanism and iterative learning mode to artificially label as few basic formulas as possible, so that the automatic derivation machine automatically derives higher-order complex formulas based on low-order simple formula templates.


Artificial intelligence methods should not only focus on tasks such as recommendation or automatic identification in the service industry, but should also be used in more meaningful scientific research areas. By establishing the formula automatic derivation machine, we can efficiently obtain meaningful results. In the case of this paper, the neural network model $\pi_{\theta}(a|s)$ is trained by constructing a first-order linear differential equation training set, which successfully derives the fission concentration equation solution of $Pm^{149}$ in reactor physics. In accordance with the formula automatic derivation machine design principle of this paper, more complex professional scientific research problems can be solved.




\section{formula multiway tree}
\subsection{multiway tree model}

In the derivation of mathematical formulas, in order for the formula to be computed by a computer, the formula must be re-expressed in a computable form. In the research of machine proof, the proof of many geometric problems is re-expressed as the related polynomial \cite{Al2017A}, and the homogeneous differential equation can also be mapped as a polynomial to calculate, for example:

$$ay'''+by''+cy'+d=y \Rightarrow ay^3+by^2+cy^1+d=y^0$$


However, the traditional polynomial expression has great limitations. For example, if you want to use a polynomial to express a complicated formula, it is not feasible. For example, such as this formula:

$$\int{ydx}+sin(y/x)+S(x)=0$$


Therefore, the formula multiway tree is proposed to re-express the formula. The formula multiway tree is constructed from top to bottom according to the symbol priority. The computational symbol is a node, and the algebraic symbols and numerical symbols are leaf nodes, so that any complex formulas can be decomposed and re-expressed in the form of a multiway tree.


The formula is re-expressed with a multiway tree, and the formula derivation process can be based on the operation and transformation algorithm of the multiway tree. In the derivation process, the structure and mathematical connotation of the formula follow the derivation rules strictly, eliminating errors and ambiguities, which is also easy to implement in programming languages.


For example, if the following formula needs to be expressed,the corresponding multiway tree of the formula is shown in FIG.1.

$$\delta = \int \frac{\bar{M_i}\bar{M_j}}{EI}dx $$
$$P(x,y,z,t)=dE/dS$$

\begin{figure}[h]
\centering
\includegraphics[scale=0.28]{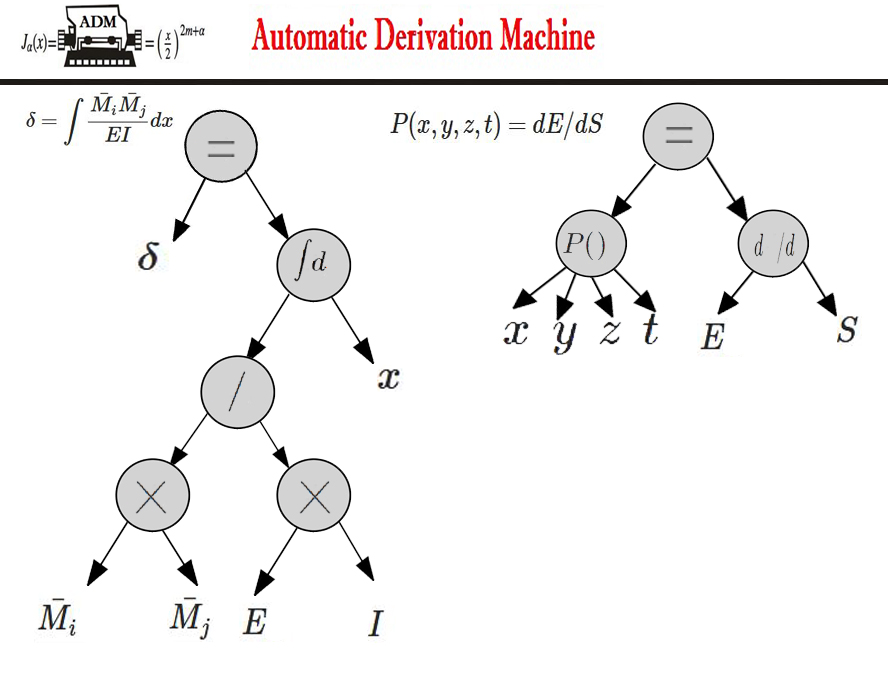}
\caption{Using a multiway tree to express the formula, each operator is a node, and algebraic symbols or other operators that belong to this operator's scope of operation belong to this node's child nodes. Algebraic symbol nodes and numerical nodes can only be leaf nodes.}
\end{figure}

\subsection{Subtree search algorithm}

In order to complete the function of iterative learning, we propose the subtree search algorithm of the formula multiway tree. The subtree search algorithm refers to when given formula multiway tree $F$ (such as $sin^2(x)+cos^2(x)=1$) and a simple template formula $T$ (such as $a+b =c$), judging whether the formula $F$ conforms to the template formula $T$, that is, whether the formula multiway tree of $F$ contains the subtree $T$.


The search algorithm is based on recursive implementation. Specifically, the subtree $T$ is matched from the root of the $F$ formula multitree, and if their symbols are the same, their child nodes would be compared, if their symbols are not the same,then start matching from the child nodes of $F$ recursively. The specific algorithm is as follows:

\begin{algorithm}[t]
\caption{subtree search algorithm $find(root,sub)$} 
\hspace*{0.02in} {\bf Input: Specific formula root node $root$; root node of the formula subtree $sub$;Specific node matching function $TreeHasSub(root,sub)$} 
\hspace*{0.02in} {\bf Output:if the template formula $sub$ is the subtree of $root$} 
\begin{algorithmic}[1]
\If{their symbols are the same: root.sym=sub.sym}
    \If{$TreeHasSub(root,sub)$} 
        return true;
    \EndIf
\EndIf
\For{$node$ in $root.subNodes$}
　　\State return find(node,sub);
\EndFor
\State return result
\end{algorithmic}
\end{algorithm}


For example, the formula $dy=adx$ before derivation below is the subtree of the formula $df(x)=sinxdx$ before derivation, and the formula $f(x)=\int{sinxdx}$ can be deduced according to the derivation pattern of $dy=adx \Rightarrow y=\int{adx}$, which is also called the template mapping method proposed later in this paper.

$$dy=adx \Rightarrow y=\int{adx} $$
$$df(x)=sinxdx \Rightarrow f(x)=\int{sinxdx} $$

\begin{figure}[H]
\centering
\includegraphics[scale=0.28]{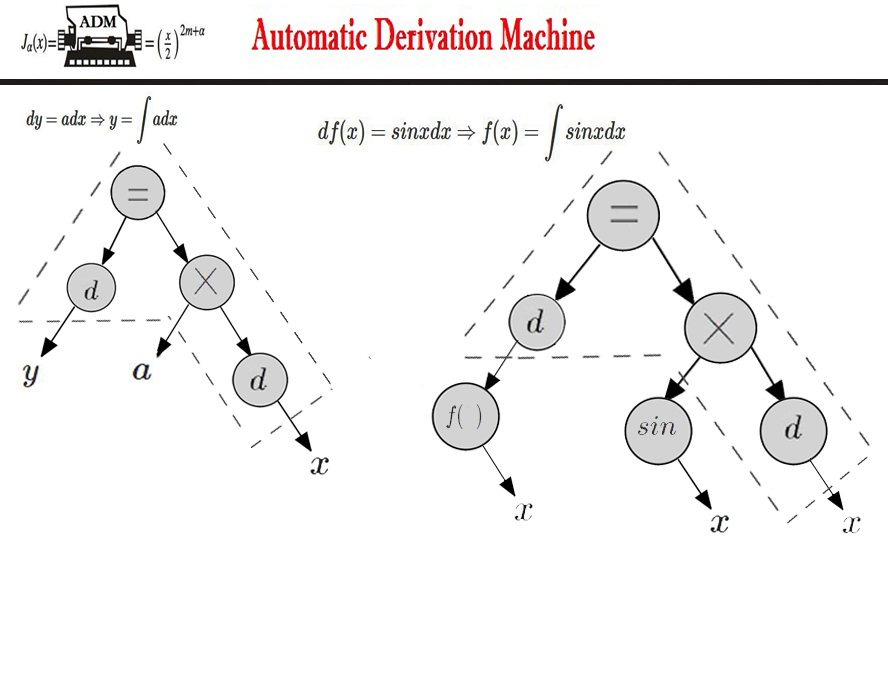}
\caption{Using formula $dy=adx$ as a subtree, performing subtree search algorithm in other formulas to find out if this subtree is included. For example, formula $df(x)=sinxdx$ contains this subtree, which is equivalent to, $df(x)=sinxdx$ is a specific case of the template formula $dy=adx$.}
\end{figure}

\subsection{Construct declaration method}

Because we need to manually build a large number of basic formulas for training in the later period,so when declaring a specific multiway tree for a formula, we need an efficient declaration method that meets people's thinking. Therefore, we propose a a multiway  tree construction method. Here, we only need to use a construct function in the programming language(C++ language is used in this article)to do the declaration of a formula. The construct function returns an instance of a formula multiway tree(in this article, the program returns the c++ point of the root node of a formula multiway tree). For example, we declare the following formula:

$$df(x)=sinxdx \Rightarrow$$
$$Equal(Der(f(Sym("x"))), Times(Sin(Sym("x")),Der(Sym("x"))))$$

\begin{figure}[H]
\centering
\includegraphics[scale=0.26]{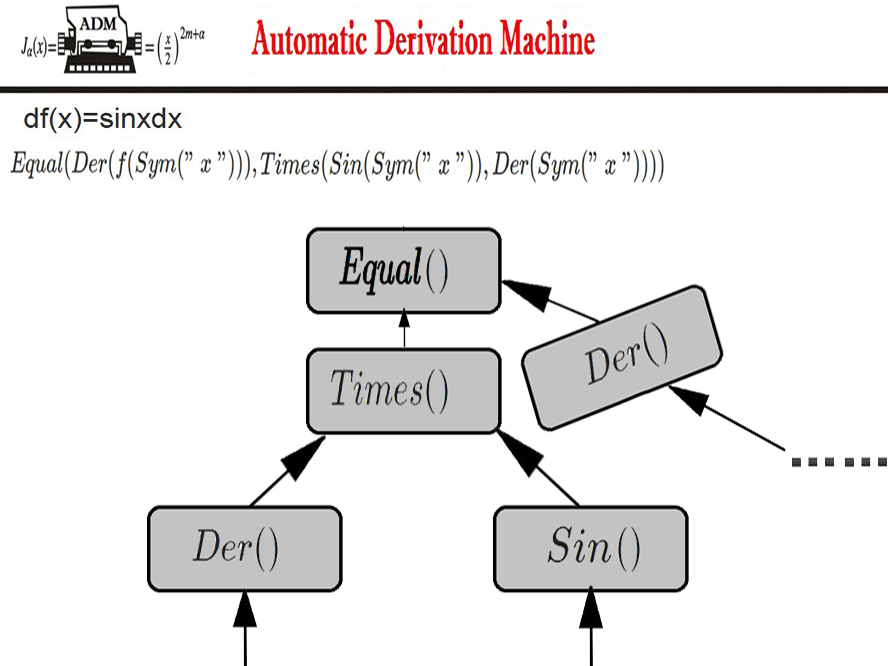}
\caption{Expressions based on multiway trees are also suitable for using code expressions. Each operator is abstracted as a functional expression, which allows top-down use of code to express arbitrarily complex formulas easily.}
\end{figure}

\section{Formula derivation implementation}
\subsection{Template mapping method}

We propose the template mapping method that allows the formula to be deduced from the current state to the next step. Specifically, for example, a one-step derivation of the formula $e^xsinx=m(x)t$:

$$e^xsinx=m(x)t \Rightarrow e^xsinx/t=m(x)$$


This can be seen as done according to the template formula $a=bc \Rightarrow a/c=b$. First,he formula $e^xsinx=m(x)t$ judges if the subtree $a=bc$ is included, and then the template $a=bc\Rightarrow a/c=b$ can be used to complete its derivation: $e^xsinx= m(x)t \Rightarrow e^xsinx/t=m(x)$.

\subsection{Derivation by template mapping method}

Template mapping can be seen as an abstraction of the specific derivation steps. It describes the transformation of the template formula $T_1$ to the template formula $T_2$, and the template formula describes the simplest formula multiway tree that conforms to this derivation pattern. The template map can be described as:

$$M(T_1)=T_2$$


Where $M(.)$ represents some kind of derived transformation, $T_1$ represents the template formula before derivation, and $T_2$ represents the template formula after derivation. For example, addition and subtraction can be described as:

$$a+b=c \Rightarrow a=c-b $$
$$M(tree(a+b=c)) = tree(a=c-b) $$


Any formula derivation can be done based on the most basic template mapping. That is, the derivation of the formula $F_1$ can be performed in a one-step derivation according to the mode of $T_1 \Rightarrow T_2$. This derivation method that depends on the template mapping can be expressed as $Trans(F_1,T_1,T_2)=Trans(F_1,T_1,M(T_1))=F_2$.  such as, for the formula $m(x)+s(y)= T(x,y)$, according to the template mapping $a+b=c \Rightarrow a=cb $, the transformation is:

$$m(x)+s(y)=T(x,y) \Rightarrow m(x)=T(x,y)-s(y)$$
$$Trans(tree(m(x)+s(y)=T(x,y)),tree(a+b=c),tree(a=c-b))$$
$$=tree(m(x)=T(x,y)-s(y))$$


This method is difficult to solve the implicit conversion with physical connotation like $\oint{\frac{m}{p}dx}\Rightarrow \oint{\frac{dx}{v}}$. The solution to this problem is to break it down into several steps to complete, and need to add auxiliary symbols to complete, specifically:


$$\oint{\frac{m}{p}dx}=S \xrightarrow{ab/c=d \Rightarrow a=dc/b} \oint{dx}=S\frac{p}{m} \xrightarrow{v=p/m}$$
$$\oint{dx}=Sv \xrightarrow{ab/c=d \Rightarrow a=dc/b} \oint{\frac{dx}{v}}=S$$

\subsection{Formula multiway tree replacement algorithm}

In order to implement the function of deducing by template mapping, we propose the replacement algorithm of the formula multiway tree. After the formula $F_1$ finds that the template formula $T_1$ is its subtree, the replacement algorithm for the corresponding node of the formula $F_1$ and the template formula $T_1$ can be performed on the template formula $T_2$ (this step is equivalent to the one in the mathematical operation: substituting another formula into the formula for calculation.), the final replaced formula $T_2$ is the deduced result $T_2$. The specific algorithm is:

\begin{algorithm}[t]
\caption{Replacement algorithm $Replace(root,sub1,sub2)$} 
\hspace*{0.02in} {\bf Input: Specific Formula A and the node $root$ that coincides with the root node of $sub1$; Template formula $sub1$ that matches formula A;Template formula $sub2$ to be transformed.} 
\hspace*{0.02in} {\bf Output:Derived result formula $sub_{copy}$} 
\begin{algorithmic}[1]
\State construct $Map = makeMap(root,sub1)$ ,find the part of the formula that corresponds to each symbol in the template formula.
\State construct $sub_{copy}$,as a copy of $sub2$.
\For{Traverse every corresponding node in $Map$} 
　　\State Replace the corresponding node on $sub_{copy}$ with the corresponding specific part of the formula;
\EndFor
\State return $sub_{copy}$
\end{algorithmic}
\end{algorithm}

\begin{figure*}[t]
\centering
\includegraphics[scale=0.18]{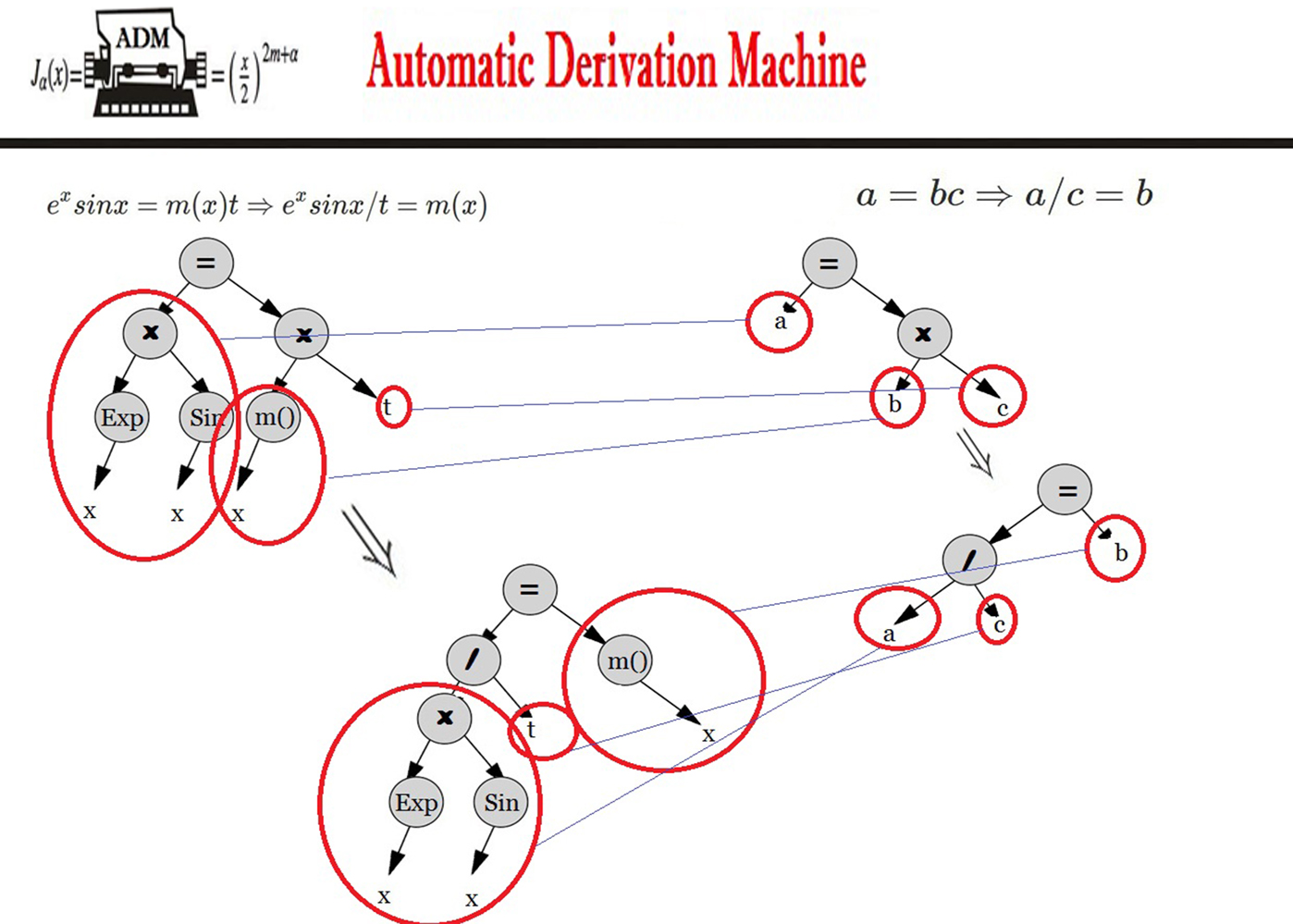}
\caption{Replace algorithm.Step 1: find the corresponding nodes of the specific formula on the subtree $sub1$ of the corresponding mode, and build the mapping relationship between them. Step 2: traverse this map and implement the replacement on the $sub2_{copy}$. Step 3: mount the $sub2_{copy}$ back to the original tree and complete a one-step derivation.}
\end{figure*}


Based on the template mapping, we propose iterative learning. The iterative learning mechanism refers to the initial basic formula derivation template $T_0=(t_01,t_02)$, where $T_0$ is a simple enough formula transformation, and $T_1=(t_11 ,t_12)$ is a formula derived from $T_0$ or a number of map templates of the same class containing $T_0$. By the mechanism like this, $T_{n-1}$ can lead to a more complex formula transformation method $T_{n}$. Such formula derivation based on iterative transformation is called iterative learning.

\begin{figure}[t]
\centering
\includegraphics[scale=0.33]{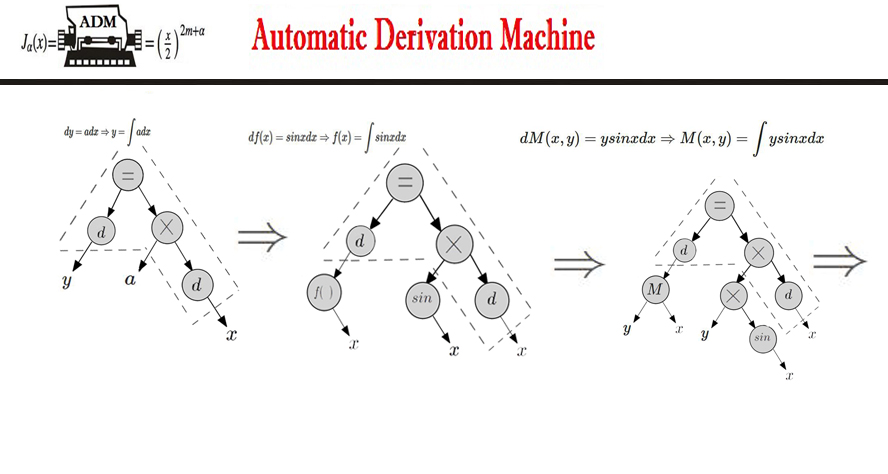}
\caption{Using the basic template derivation formulas, the derived formulas continue to be used as templates for derivation, which is called iterative learning. Iterative learning can lead to complex formulas starting from the simplest formulas transformation.}
\end{figure}

\section{Formula multiway tree encoding}

\subsection{Benefits of feature encoding}

Although the formula multiway tree can achieve derivation by searching for subtrees and replacement algorithms, formulas using multiway tree expressions cannot be input into the learner directly for training fitting and error metrics. When we need to measure the similarity of the following two formulas, the tree model is not intuitive:

$$te^x+mcosx,te^{-x}-asinx$$

Therefore, we propose coding algorithm to encode the formula multiway tree and transform the formula multiway tree into the corresponding feature vectors. Which has three advantages:

\begin{itemize}
\item Each formula has a unique identifier.
\item The similarity of the formula can be measured, the similar formula multiway trees have a smaller spatial distance.
\item The formula multiway tree can be transformed into a feature vector to train the learner.
\end{itemize}

\subsection{Encoding method}

The specific encoding method is to assign an integer tag to each operator that appears in the formulas, do the similar Breadth-first traversal to the formula multiway tree from the top, and put the integer encoding of its nodes and child nodes into the feature vector. The specific process is as follows:

\begin{algorithm}
\caption{Formula multiway tree encoding} 
\hspace*{0.02in} {\bf Input:formula multiway tree root node $root$; } 
\hspace*{0.02in} {\bf Globle:Maximum encoding length $L_{max}$;Global encoding sequence $V$;Global encoding set ${N_i}$ }
\hspace*{0.02in} {\bf Output:feature vector $V$} 
\begin{algorithmic}[1]
\State $encode(root)$
\State judge $L_V<L_{max}$
\State clear temporary sequence $S$:$S=[]$;
\State Insert the root node code at the end of the temporary sequence: $S.push(N_{root})$
\State \For $node \in root.SubNodes$:$S.push(N_{node})$\EndFor
\State $V.push(S)$;
\State \For $node \in root.SubNodes$:if $node.type$ is not "Sym":$encode(node)$;\EndFor
\State finally, insert $L_{max}-L_{V}$ 0 after $V$；
\end{algorithmic}
\end{algorithm}

\begin{figure}[H]
\centering
\includegraphics[scale=0.28]{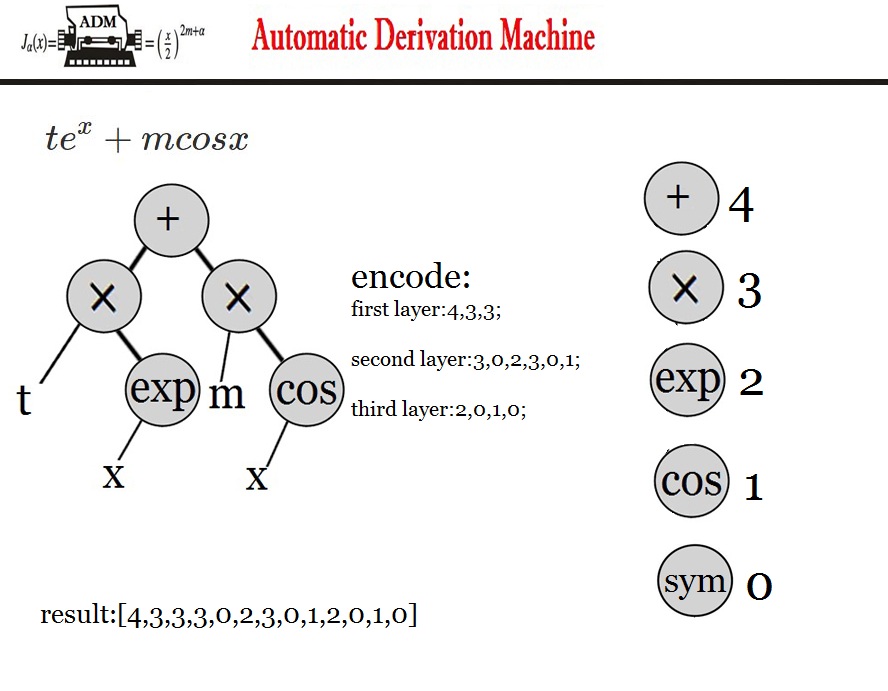}
\caption{The multiway tree encoding algorithm is a top-down, breadth-first algorithm that assigns integer coding to all symbols firstly. Both the algebraic and numeric symbols are encoded as 0. Then the sub-layers are encoded from left to right.}
\end{figure}

\subsection{Difference measure}

After obtaining the eigenvectors of the two formula multiway trees, we need to define the calculation method for measuring the difference $dis(tree_1, tree_2)$ between them. First of all, we have ensured that all the eigenvectors are of equal dimensions, and then compare each bit of them. If they are not equal, the comparison result is 1, otherwise, the comparison result is 0. Finally, the comparison result is the difference between the two formulas. For example, for the difference value calculation of formulas below:

$$dis(tree(te^x+mcosx),tree(te^{-x}-asinx))$$
$$=dis([4,3,3,3,0,2,3,0,1,2,0,1,0],$$
$$[5,3,3,3,0,2,3,0,6,2,3,6,0,3,0,0])$$
$$=4$$


Defining the formula multiway tree encoding method and differential value calculation method is of great significance to the follow-up learner training and prediction function. The encoding method can extract the features of the formula multiway tree, and convert it into a feature vector which can be calculated using the linear algebra method. The difference value calculation method can be used to complete the error convergence of the neural networks model.

\begin{figure}[H]
\centering
\includegraphics[scale=0.28]{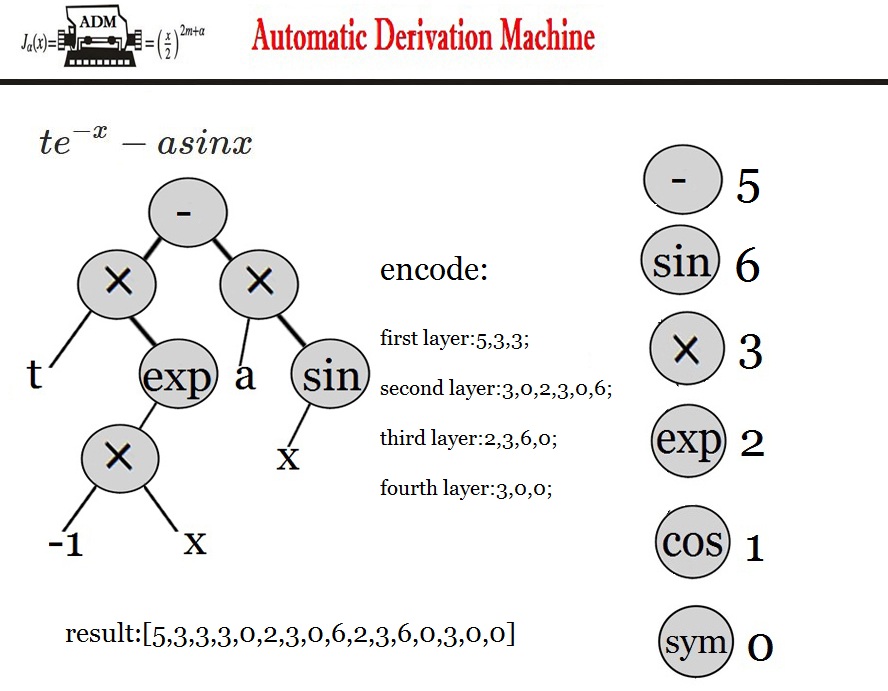}
\caption{Formula multiway tree encoding of another tree.}
\end{figure}

\section{Learning method of formula derivation}
\subsection{Reinforcement learning mode}

We choose the reinforcement learning method to train the formula derivation machine. The reinforcement learning mode can be described as extracting an enviroment from the task to be completed, abstracting the state, action, and instantaneous reward received from performing the action \cite{Sutton1995Generalization}. The key elements of reinforcement learning are: environment, reward, action, and state. With these elements, a reinforcement learning model can be established. The problem of reinforcement learning is to obtain an optimal strategy for a specific problem, so that the reward obtained under this strategy is maximized. The so-called strategy is actually a series of actions, that is, sequential data.


We use the Q-learning learning model \cite{Mnih2016Asynchronous} in this paper. Q-learning's learning model needs to update a table named Q-table continuously. In this table, $Q(s,a)$ represents the action $a$ we choose at the state $s$, and Q-learning's learning process is to update every value in this table, that is, the expected earnings when taking action $a$ under any state $s$. The specific update method is:

$$Q(s,a)=R(s,a)+\gamma MAX_{a'}Q(s',a') $$


Through continuous updating by training data, a converged Q-table can be finally used to select the action $a$ with the largest expectation of revenue for the specific state $s$. The action can be thought of as a transformation of the state $s$ into $s'$, which can be expressed as $a(s)=s'$.

\begin{figure}[H]
\centering
\includegraphics[scale=0.3]{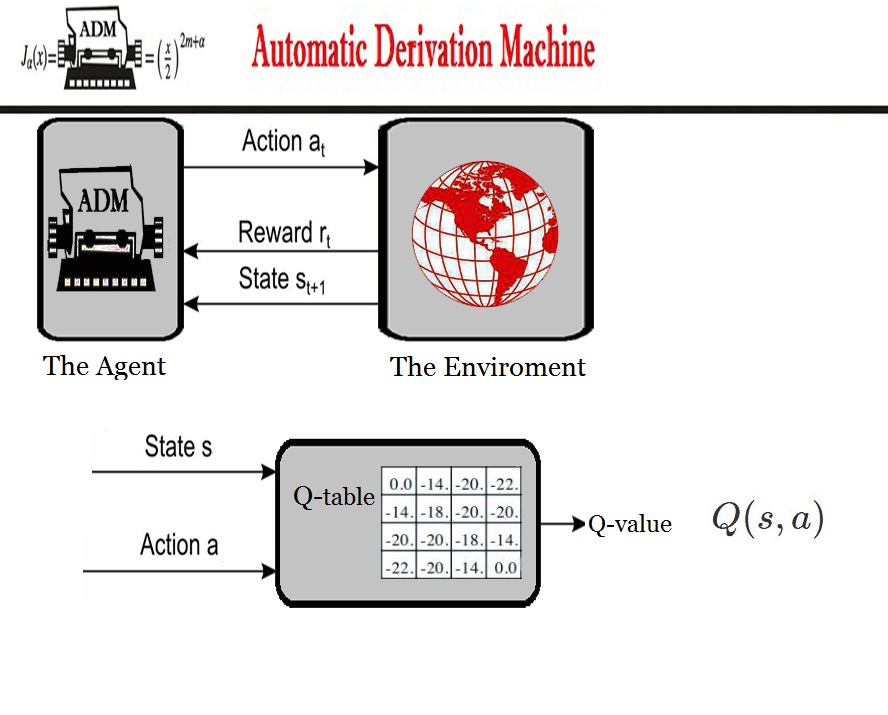}
\caption{Q-learning is a state-based decision model, where each pair of $(s,a)$ has a corresponding Q-value quantification in the Q-table to take the plausibility of acting $a$ under the state $s$, and Q-table is the target for learning.}
\end{figure}

\subsection{Reinforcement learning mode for automatic formula derivation}

When the reinforcement learning mode is used to solve the problem of automatic formula derivation, the state set of the environment corresponds to the form of the formula multiway tree at different stages of derivation \cite{Farquhar2018TreeQN}, and the agent-selectable action set corresponds to all the basic templates that can be selected when formulas deduced. The action can be taken as the template mapping $(T_{at}, T_{bt})$ converts $F_t$ to the formula $F_{t+1}$, which is the conversion method $R(F_t,(T_{at},T_{bt}))=F_{t+1}$. Specifically it can be expressed as:

$$S={s1,s2...}={F1,F2...}$$

$$A={a1,a2...}={(T_{a1},T_{b1}),(T_{a2},T_{b2})...}$$


Therefore, the automatic derivation machine needs to calculate the decision probability $\pi_{\theta}(a|s)$ at each step, that is, select the action $a$ with the largest expected return (the one that is closest to the problem target) according to the existing state $s$. Then take the subtree search and replace algorithm to achieve a one-step formula derivation. The specific form of this probability:

$$\pi_{\theta}(a|s)=\pi_{\theta}((T_{at},T_{bt})|F_t)$$


The neural network model and gradient descent method are used to realize the calculation of $\pi_{\theta}(a|s)$(in fact, any other learning method can be used here to find this probability, and the accuracy may be better than neural network). The eigenvector of the formula multiway tree $F_t$ is taken as input, and the output is the probability of the selectable action $a$.

\begin{figure}[H]
\centering
\includegraphics[scale=0.3]{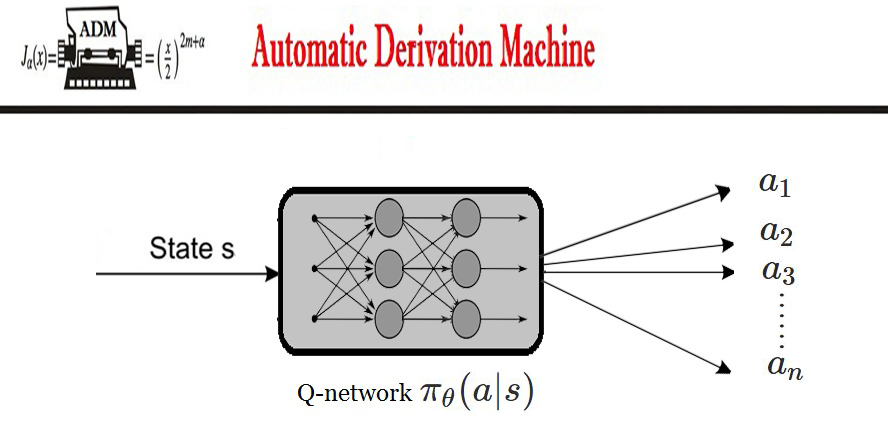}
\caption{Using neural networks to learn the state-based decision map $\pi_{\theta}(a|s)$, the output is the probability distribution in the entire action space. The selected action is the action corresponding to the output probability with maximum value.}
\end{figure}

\subsection{Training of formula derivation}

It is necessary to construct a training set about formula derivation before learning begins. The input element in each round of training is a formula $F_t$ at a certain step. The output element of the training set is the optimal transformation selection $(T_{at},T_{bt})$. That is, $(X,y)={F_t,(T_{at},T_{bt})}$.


Therefore, a complete formula derivation example can be decomposed into multiple derivation steps. Each step is a training sample, so a complete formula derivation can be converted into multiple training samples. A complete formula inference example can be expressed as:

$$F_0 \xrightarrow{(T_{a0},T_{b0})} F_1 \xrightarrow{...} F_t \xrightarrow{(T_{at},T_{bt})} F_{t+1} \xrightarrow{...}$$


The derivation of the solution to a certain physical equation velocity can be expressed as:

$$mv^2/2+E=Q \xrightarrow{(a+b=c,a=c-b)} mv^2/2=Q-E \xrightarrow{...} $$
$$v^2=2(Q-E)/m \xrightarrow{(a^2=b,a=\sqrt b)} v=\sqrt{ 2(Q-E)/m} \xrightarrow{...}$$


The specific learning error improvement uses a gradient descent method:

$$\theta \leftarrow \theta - \epsilon \delta L(\pi_{\theta}(a|s),a^*)$$


In addition, in the training data, the probability of transformation method $a^*=(T_{at}, T_{bt})$ needs to be converted into one-hot vector, which is $a^*=(0,0,0... 1...0)$.

\begin{figure}[H]
\centering
\includegraphics[scale=0.3]{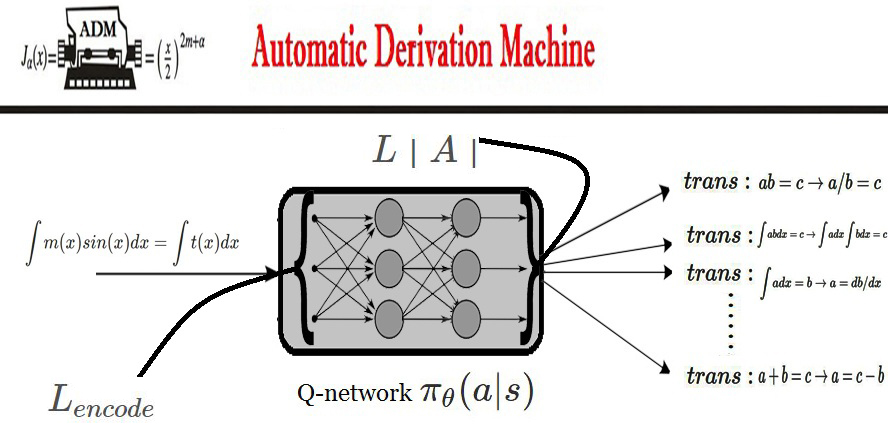}
\caption{Use neural networks to learn $\pi_{\theta}(a|s)$. The result of the output is the probability of taking each formula transformation.}
\end{figure}

\section{Problem examples}

Now we use the above method to solve a practical problem, solving the concentration equation of $Pm^{149}$ in reactor physics, which is a first-order linear differential equation:

$$dN_{Pm}(t)/dt = \gamma \sum \phi - \lambda N_{Pm}(t) $$


Firstly we need to construct the formula derivation training set for the first-order linear differential equation. The form of the first-order linear differential equation is like $dy/dx + P(x)y = Q(x)$. We hope that the auto-derivation can not only solve the concentration equation, it can also solve all first-order linear differential equations. So $P(x)$ or $Q(x)$ is equivalent to a collection of functions, that is:

$${P(x)_i}={a,ax,ax^2...ax^n...e^x,sinx...} $$
$${Q(x)_i}={a,ax,ax^2...ax^n...e^x,sinx...} $$


The training set $D={(X_i,y_i)}$ is constructed using equation inference examples containing these functions. The input element of the training set is a formula $X_i=F_t$ at a certain step, and the output element of the training set is the best transformation chooses $y_i=(T_{at}, T_{bt})$, and finally we get the conditional probability $\pi_{\theta}(a|s)=\pi_{\ Theta}(y|X)$ to make a derivation decision.


In the derivation process, each deduction will evaluate the next template selection according to the value of $\pi_{\theta}(a|s)$, and then transform the formula multiway tree according to the subtree matching and replacement algorithm, and then proceed to the next step until the final goal formula multiway tree is reached.

$$dN_{Pm}(t)/dt = \gamma \sum \phi - \lambda N_{Pm}(t)    \xrightarrow{(a/b=c,a=cb)}$$
$$dN_{Pm}(t)=(\gamma \sum \phi - \lambda N_{Pm}(t))dt  \xrightarrow{...}  $$
$$\int{1/(\gamma \sum \phi-\lambda N_{Pm}(t))}=\int{dt} \xrightarrow{(S=\int{dt},S=t)}  $$
$$\int{1/(\gamma \sum \phi-\lambda N_{Pm}(t))}=t  \xrightarrow{...}  $$




\begin{table*}
\caption{Encoding of operators appearing in formulas in training data.}\label{tab1}
\begin{tabular}{|l|l|l|}
\hline
Class &  Encoding of operators \\
\hline
Compute symbols & {$Sym/Num,+,-,\times,=,\int d,\sum,\phi,/,\sqrt,d, ln(), exp(), d/d$}  \\
Encoding number & 0,1,2,3,4,5,6,7,8,9,10,11,12,13,14,15,16 \\
\hline
\end{tabular}
\end{table*}


\section{Discuss}

In order to obtain an automatic derivation machine that can derive mathematical formulas for various professional scientific research fields, we express the mathematical formulas in a multiway tree firstly and propose a subtree search algorithm and replacement algorithm for the formula multiway trees. Implementing the formula's iterative learning function, we defined a method of formula derivation based on a simple formula mapping template. That is, only the simplest basic formula transformation set is given, and the subtree search algorithm and replacement algorithm of the formula multiway tree are used to make complex formulas follow the simple formula's transformation rules. In order to measure the difference of the formulas and facilitate the learner learning the formula's connotation, we propose an algorithm for encoding the formula multiway tree, transform the formula multiway tree into a feature vector. Finally, the neural network is constructed using the reinforcement learning model, and the gradient descent method is used to train the network to make the correct transformation decision according to the current formula state.


Although it is possible to deduce the correct answer, the training process is still very complicated because of the large amount data of neural network training required. Constructing a training set by man is a relatively large project, and the specific formula is more complicated such as a differential equation, so the requirements for training data producers are also higher than the requirements of the general machine learning data producers. Therefore, we should consider using smarter methods to generate formula data. For example, if writing a formula for training formulas according to certain rules, making training data will be more efficiently and accurately.


The encoding method of the formula multiway tree is still flawed, although the encoding method in this paper can measure the similarity of the formula structure, it can not measure the similarity between the calculated symbols. The encoding method of this paper will determine that the difference between $+$ and $E(x,y)$ and difference between $+$ and $-$ is equivalent, but in fact, for the specific problem, the difference and connotation between the symbols are not equivalent. Learning-based encoding methods like word2vec \cite{Mikolov2013Distributed} will improve the above problem.


The formula multiway tree model and formula automatic derivation machine is a preliminary study for the automation and intelligence of scientific research, but it proposes a relatively efficient framework for re-expressing the abstract mathematical formulas of different scientific research fields and establishing a general decision model to further derivation. Such a learning framework can be used for more complex and difficult scientific research from cross-domains, what needs to be done manually is to set basic rules and patterns, then using automatic derivation can be more efficient than artificially iterative learning and derivation. We hope that the automatic derivation machine will achieve meaningful derivation in specific scientific research fields.

\bibliographystyle{ACM-Reference-Format}
\bibliography{sample-bibliography}

\end{document}